\RequirePackage{fix-cm}
\documentclass[smallextended]{svjour3}       
\smartqed  
\usepackage{url}

\usepackage{graphicx}
\usepackage[usenames,dvipsnames]{xcolor}
\usepackage{caption, subcaption}
\usepackage{booktabs}
\usepackage{mathrsfs}
\usepackage{microtype}
\usepackage{setspace}

\usepackage{placeins}
\usepackage{multirow}
\usepackage{rotating}
\usepackage[normalem]{ulem}
\usepackage{amsfonts}       
\usepackage{nicefrac}       

\usepackage[hyperfootnotes=false]{hyperref}
\hypersetup{
	colorlinks=true,
	linkcolor=blue!30!black,
	citecolor=green!30!black,
	linkbordercolor=blue!30!black,
	citebordercolor=green!30!black,
	pdfborderstyle={/S/U/W 0.3}
}

\usepackage{macros}

\newcommand{\Dictionary}{ \mathscr{D} }

\newcommand{\simplex}{ \mathbb{P} }
\newcommand{\bld}[1]{ \textbf{\textcolor{gray!60!black}{#1}} }

\newcommand{\besth}[1]{ \textbf{\textcolor{gray!50!green}{#1}} }

\begin{document}

\title{Natural Alpha Embeddings
}

\author{Riccardo Volpi
\and
        Luigi Malag\`{o} 
}


\institute{Riccardo Volpi\href{https://orcid.org/0000-0003-4485-9573}{\includegraphics[scale=0.5]{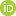}}, Luigi Malag\`{o} \at
              Romanian Institute of Science and Technology \\
	        Cluj-Napoca, Romania\\
              \email{\{volpi,malago\}@rist.ro}           
}

\date{Received: date / Accepted: date}

\maketitle

\begin{abstract}
Learning an embedding for a large collection of items is a popular approach to overcome the computational limitations associated to one-hot encodings. The aim of item embedding is to learn a low dimensional space for the representations, able to capture with its geometry relevant features or relationships for the data at hand. This can be achieved for example by exploiting adjacencies among items in large sets of unlabelled data. In this paper we interpret in an Information Geometric framework the item embeddings obtained from conditional models. By exploiting the $\alpha$-geometry of the exponential family, first introduced by Amari, we introduce a family of natural $\alpha$-embeddings represented by vectors in the tangent space of the probability simplex, which includes as a special case standard approaches available in the literature.
A typical example is given by word embeddings, commonly used in natural language processing, such as Word2Vec and GloVe. In our analysis, we show how the $\alpha$-deformation parameter can impact on standard evaluation tasks.
\end{abstract}

\section{Introduction}

Item embedding is a collective name for a set of techniques extracting meaningful representations from a huge amount of unlabelled data, by exploiting the complex network of relationships among the set of items in a dictionary.
Studying the geometry of the learned embedding space is fundamental to understand the kind of information which has been extracted and how it has been organized~\cite{michel_does_2017,coenen_visualizing_2019,hewitt_structural_2019}.
Popular applications of interest range from recommendation systems~\cite{koren_matrix_2009,barkan_item2vec:_2016,krishnamurthy_learning_2016,zhao_learning_2018} to approximate similarity based search and information retrieval for a wide variety of tasks~\cite{sugawara_approximately_2016,wu_starspace:_2018}.
In this paper we will study item embeddings with a particular focus on natural language processing, which allows us to provide intuitive examples based on common understanding.

Natural Language Processing (NLP) is a branch of machine learning which deals with the design of algorithms to effectively process natural language corpora. The one-hot encoding is a common representation for the words in a dictionary, each word is assigned to a different direction in the space, whose total dimension corresponds to the size of the dictionary $n=|\Dictionary|$, where $\Dictionary=\left\{ x\in \{0,1\}^n \, | \, \sum_{i} x_i =1 \right\}$. Such sparse representation has the characteristic that all words $w\in\Dictionary$ are equidistant between each other and no metric is implicitly defined a priori on the space. However, a practical issue in NLP arises in presence of a very large dictionary $\Dictionary$, for instance, training a neural network taking in input a sequence of $n$-dimensional vectors, becomes quickly prohibitive even for relatively limited language domains. The aim of item embedding, or word embedding in this specific case, is to project the one-hot encoding vectors onto a lower-dimensional space, mapping $w\in \Dictionary$ to $u\in \mathbb{R}^d$, with $d \ll n$.
Several probabilistic models can be designed and subsequently trained to learn these compact representations. Notice that the training of the model at this stage is completely unsupervised, guiding the representations to leverage the huge amount of unlabelled text available.

Neural network language models~\cite{rumelhart_learning_1986,bengio_neural_2003} introduce the idea of using the internal representation of a neural network to construct the word embedding. In particular Bengio et al.~\cite{bengio_neural_2003} were among the first ones to propose the use of a neural network to predict the probability of the next word given the previous $N$ ones ($N$-gram model). A $n\times d$ matrix $C$ is used to project the one-hot encoding of the previous words onto a linear space of dimension $d$. Then a neural network is used to generate the probabilities of the next word, given the concatenation of the projection of the $N$ previous ones. The words representations and the weights of the neural network are learned simultaneously, in order to obtain a matrix $C$ able to project similar words close to each other. Mikolov et al.~\cite{mikolov_recurrent_2010} are the first ones introducing a recurrent language model, using a Recurrent Neural Network (RNN) to model the vector representations in the embedded space $\mathbb{R}^d$. Such approach has the advantage of using less weights compared to an N-gram approach, while still potentially being able to learn quality word embeddings depending on the previous words (given the limitations of learning long-time dependencies with vanilla RNNs, cf.~\cite{bengio1994learning}). 

It is a well known, and at first surprising fact, that syntactic and semantic analogies between words (e.g., $king:man=queen:woman$) translate into vectorial relationships between the respective word vectors learned by the embedding, e.g.,
($u_{king} - u_{man}~\approx u_{queen} - u_{woman}$)~\cite{mikolov_linguistic_2013,pennington_glove:_2014,lee2015linear,levy_neural_2014,arora_rand-walk:_2016,arora_linear_2016}.
Even relatively simple models have been shown to work well at capturing syntactic and semantic analogies, in particular we can mention the Skip-Gram (SG)~\cite{mikolov_efficient_2013,mikolov_distributed_2013} and the Continuous Bag Of Words model (CBOW)~\cite{mikolov_efficient_2013}. In such models the training task is to predict a word from its context or vice versa. SG tries to predict the words in the context from the central word, while CBOW makes a sum of the representations of all the vectors of the context and tries to predict the central word. A slightly different approach based on global statistics of the corpus is given by GloVe (Global Vectors), introduced by Pennington et al.~\cite{pennington_glove:_2014}. The novelty of this approach is that it is learning directly from the counts of the co-occurrences, thus it does not need to iterate over the corpus during training as SG. This is particularly advantageous for big corpora in which the dimension of the corpus is much bigger than the dimension of the global matrix of co-occurrences.

There have been numerous works investigating the origin of the linear  structure of word analogies in the embedding space. Pennington et al.~give a clear intuitive explanation in their paper on GloVe~\cite{pennington_glove:_2014}.
More recently Arora et al.~\cite{arora_rand-walk:_2016} formalized this further by introducing a Hidden Markov Model for text generation and assuming that word embedding vectors are isotropically distributed in space after learning. Nevertheless, how well information is actually encoded in the space of word embeddings seems to not be completely understood yet. The embedding of a particular word is based on the co-occurrences within the context, i.e., words with similar context tend to be projected nearby in the embedded space. This makes unclear if opposite words will be near to each other in the embedded space (e.g., hot and cold) and in general how well words will be spaced. An additional motivation for the need to study the word embedding distances is that, in evaluating accuracies on analogies $a:b=c:\,?$, is a common practice to remove the three query words $a$, $b$, $c$ from the set of possible returned results, otherwise the answer is often one of these three query words. A further problem related to the expressivity of the embedding space is word polisemy~\cite{arora_linear_2016}, that is still object of investigations. More recent studies aim to deploy different models, based on transformers (suitable for capturing long-term dependencies), to solve some of these issues and increase the overall performances of the learned embeddings~\cite{radford_improving_2018,devlin_bert:_2018,yang_xlnet:_2019,hewitt_structural_2019,coenen_visualizing_2019}. This opens up to a plethora of language models and their associated pretraining strategies for the respective word embeddings. It is clear that understanding deeply the meaning of distances and directions in the word embedding space in the different models is of key importance for numerous NLP tasks using word embedding as a base block.

We want to stress how all the different strategies enlisted so far have in common the parametrization of one or more discrete probability distributions over the dictionary, i.e., the identification of a point in a probability Simplex. Our aim is to build a bridge between the geometrical view of Information Geometry and the probabilistic models applied in the literature.
In this study, we will focus our analysis on standard word embeddings models based on skip-gram conditional probability model. Using notions of Information Geometry we will interpret the embedding as a vector in the tangent space of the manifold. Next, by exploiting the notion of $\alpha$-connection for dually flat statistical models, first introduced by Amari~\cite{amari1980theory,amari1982differential,amari1982geometrical,nagaoka1982differential,amari:85}, see also~\cite{lauritzen1987statistical}, we define a family of $\alpha$-embedding, which depends on the choice of the connection.

The use of Riemannian methods have been explored previously in the literature of NLP. Lebanon~\cite{lebanon2006metric} has been one of the first authors to propose to learn a  distance metric over the input space using a framework based on Information Geometry, with applications to text document classification. More recent applications of Riemannian optimization algorithms can be found in the work of Fonarev et al.~\cite{fonarev2017riemannian}, who proposed the use of Riemannian methods to optimize the Skip-Gram Negative Sampling objective function over the manifold of required low-rank matrices, and Nickel and Kiela~\cite{nickel2017poincare} who introduced an approach to learn hierarchical representations of symbolic data by embedding them into hyperbolic space, with applications to word embedding. More recently Jawanpuria et al.~\cite{jawanpuria2019learning} proposed a geometric framework to learn bilingual mappings given monolingual embeddings and a bilingual dictionary, where the mapping problem is framed as a classification problem on smooth Riemannian manifolds. 
The notions of $\alpha$-divergence~\cite{amari:85,amari:87dual,amari|nagaoka:2000,amari_information_2016}, has also been already exploited in several applications, for example also in the social siences~\cite{ichimori2011rounding,wada2012divisor}.

\section{Conditional Models and the Embeddings Structure}
\label{sec:exp_model}
Let $\Dictionary$ be a dictionary of cardinality $n$, presented by a one-hot encoding, i.e., $\Dictionary = \{\chi \in  \{0,1\}^n : \sum_{i=1}^n \chi_i = 1\}$. Such encoding does not impose any structure on the space of the representations, since all words are equidistant, but also it is not a practical representation for large dictionaries, and dimensionality reduction techniques are usually required.
A word embedding is a mapping from $\Dictionary$ to a lower-dimensional vector space $\reals^d$, with $d \ll n$.
One of the simplest, but still effective, models in the literature is the Skip-Gram conditional model~\cite{mikolov_distributed_2013,pennington_glove:_2014}, modelling the conditional probability distribution of the words in the context given the central word. Incidentally, the idea behind this model can be also found in a famous quote by Firth, \textit{``you shall know a word by the company it keeps''}~\cite{firth_synopsis_1957}. For each word $w \in \Dictionary$, let us fix a window $W$ around $w$, and let the context of $w$ be the set of words $\chi \in W$. The skip-gram model~\cite{mikolov_efficient_2013,pennington_glove:_2014} associates to each word $w$ a conditional probability distribution $p(\chi | w)$ with $\chi \in \Dictionary$, expressed by
\begin{equation}
\label{eq:condprobmodel}
p(\chi | w) = \frac{\exp(u_w^\trasp v_\chi)}{Z_w}~, \qquad \text{with\ }  Z_w = \sum_{\chi' \in \mathscr D} \exp(u_w^\trasp v_{\chi'})~.
\end{equation}
Notice that such model assigns to each word two vectors $u,v \in \reals^d$, which are used in case the word in question is the central word or a word of the context, respectively. The vectors $u_w, v_w$ for $w \in \Dictionary$ form two $n \times d$ projection matrices $U, V$,
with rows given by $u_w$ and $v_\chi$. The matrices $U$ and $V$ can be learned from the data by maximum likelihood estimation, with different objective losses: the likelihood of the word couples observed in the corpus (word2vec~\cite{mikolov_efficient_2013,mikolov_distributed_2013}), or using a stochastic matrix factorization approach, like in GloVe~\cite{pennington_glove:_2014}.

It has been demonstrated in the literature~\cite{levy_neural_2014} that word2vec Skip-Gram with negative sampling~\cite{mikolov_distributed_2013} is indeed equivalent to a matrix factorization \`a la GloVe. Then without loss of generality we will mainly focus on GloVe as a training methodology, since it is more computationally convenient for large corpora. The matrices $U$ and $V$ can be learned from the data by optimizing
 \begin{displaymath}
 \mathcal L(U,V | \mathscr C) = \sum_{(w,\chi) \in \Dictionary\times\Dictionary}  f(C_{w\chi}) \left( u_w^\trasp v_\chi + b_w + \tilde{b}_\chi - \ln C_{w\chi} \right)^2 ~,
 \end{displaymath}
 where $C$ is the matrix counting all the co-occurrences of couples of words $(w,\chi)$ in a corpus $\mathscr C$. The function $f$ weights the error in the matrix factorization, depending on the frequency of the couple of words in question. A typical choice is
 \begin{equation}
 f(x) =
 \begin{cases}
 \left( x/x_{max} \right)^\frac{3}{4} \qquad & x<x_{max} \\
 1 \qquad & x \ge x_{max} \;,
 \end{cases}
 \end{equation}
where $x_{max}$ is a cutoff, usually fixed to 100, cf.~\cite{pennington_glove:_2014}.
The conditional model in Eq.~\eqref{eq:condprobmodel} corresponds to the exponential family
\begin{displaymath}
\label{exp-family}
	p(\chi | w) = \frac{\exp( u_w^\trasp V^T \chi )}{Z_w}~,
\end{displaymath}
where $u_w, V$ are the parameters of the model. The family is not written in canonical form, since the vector of natural parameters corresponds to $u^\trasp V^\trasp$, with sufficient statistics $T_k(\chi) = \chi_k$, $k \in 1, \dots, n$, see for instance Section 3.4 from~\cite{casella|berger:2001}.
In the following we adopt a different perspective. We consider the matrix $V$ fixed after the inference process, so that the exponential family can be written in canonical form with natural parameters $u_w \in \reals^d$ and sufficient statistics $T_k(\chi) = (V^\trasp \chi)_k = (v_\chi)_k$, for $k = 1, \dots, d$, cf.~\cite{rudolph|maja|francisco|mandt|blei}.
The vector $u$ defines a family of conditional probability distributions and each $p(\chi | w)$, for $w \in \Dictionary$ corresponds to a different distribution with parameters $u_w$ in the exponential family identified by a fixed $V$. By conditioning over different $w$, we obtain different probability distributions which all belong to the same exponential family. Let us stop for a moment and define a bit of notation at this point. We will refer to the rows of the matrix $V$ as $v_\chi$ or $V^\chi$, and to its columns as $V_k$. In this way the following expressions define the same element $(v_\chi)_k = V_k^\chi$. We will use one or the other notation thorough the paper, according to convenience.

Once the embeddings $U$ and $V$ have been learned, a typical task of interest consists in evaluating similarities between words. We refer to the proper literature for the different measures proposed~\cite{bullinaria_extracting_2007,mikolov_distributed_2013,pennington_glove:_2014,mu_all-but--top:_2017}.
Another task of interest is the evaluation of analogies. Starting from an analogy of the form ${a:b=c:d}$, Mikolov et al.~\cite{mikolov_linguistic_2013} showed how it can be efficiently solved for one of his arguments, for instance $c$, by
\begin{equation}
	\label{eq:argminrelvectors}
	\arg \min_c || u_a - u_b - u_c + u_d||^2\;.
	\end{equation}
There have been several attempts to interpret such linear behavior, see for example~\cite{pennington_glove:_2014,levy_neural_2014} and~\cite{arora_rand-walk:_2016}. In the following we provide an intuitive explanation starting from the argument of Pennington et al.~\cite{pennington_glove:_2014}, according to which, for the words satisfying an analogy, the relationship between the contexts of the word $a$ and the word $b$ is the same as the relationship which intercurs between the contexts of the words $c$ and $d$. Solving an analogy then corresponds to finding $c$ such that
	\begin{equation}
	\label{eq:argminrelprobs}
	\arg \min_c \sum_{\chi\in \Dictionary} \left( \ln \frac{p(\chi|a)}{p(\chi|b)} - \ln \frac{p(\chi|c)}{p(\chi|d)} \right)^2~,
	\end{equation}
i.e., by minimizing the average over all possible words of the context $\chi$ of difference between the ratios of probabilities. We observe that under two hypothesis, namely the isotropy of the covariance matrix associated to the row vectors of $V$, and the ``stability'' of $Z_w$ in Eq.~\eqref{eq:condprobmodel} with respect to $w$ (i.e., $Z_a \simeq Z_b$ for any $a,b\in \Dictionary$), then Eqs.~\eqref{eq:argminrelprobs} and~\eqref{eq:argminrelvectors} are equivalent. Indeed using the isotropy of the $v_\chi$ we can write
\begin{equation}
\label{eq:argminrel}
\begin{split}
\arg & \min_c || u_a - u_b - u_c + u_d||^2 \\
&= \arg \min_c \sum_{\chi \in \Dictionary} (u_a - u_b - u_c + u_d)^\trasp (v_\chi v_\chi^\trasp) (u_a - u_b - u_c + u_d) \\
&= \arg \min_c \sum_{\chi\in \Dictionary} \left( \ln \frac{p(\chi|a)}{p(\chi|b)} - \ln \frac{p(\chi|c)}{p(\chi|d)} + \ln \frac{Z_a}{Z_b} - \ln \frac{Z_c}{Z_d} \right)^2 \;,
\end{split}
\end{equation}
which, by using the stability of the $Z$s reduces to Eq.~\eqref{eq:argminrelprobs}.

The hypothesis of isotropy and stability of the $Z_w$ have been discussed in~\cite{arora_rand-walk:_2016} and in particular the stability of $Z_w$ has been experimentally verified for 4 different word embedding objectives, namely Squared Norm, GloVe, CBOW and SG, see also~\cite{mu_all-but--top:_2017}.
Eq.~\eqref{eq:argminrel} is of particular interest for this paper, since this formula will be generalized in Section~\ref{sec:geometry_embedding}. For the moment let us just notice that, by considering the columns of $V$ as centered sufficient statistics (i.e., in case the rows of $V$ have zero mean), $V^\trasp V$ is proportional to the Fisher information matrix in the tangent space of the uniform distribution. This statement will be made more precise in the next sections (Sec.~\ref{sec:geometry_embedding}).

\section{Over-parametrization of the Simplex and Mapping to the Sphere}
\label{sec:subsimplsphere}
The exponential family of Eq.~\eqref{eq:condprobmodel} represents in reality a submodel of the simplex. The full $n-1$ dimensional simplex, embedded in $\reals^n$, is the set $\mathbb{P}^{n-1}=\{\mu \in \reals^n | \sum_{i} \mu_i =1,\, \mu_i>0 \; \forall i\}$. As usually happens in the machine learning community, an over-parametrization of the simplex is commonly used, through a function called softmax. This consist in taking $\theta \in \reals^n$ and mapping this to the point $F(\theta) = \mu\in \mathbb{P}^{n-1}$ such that 
\begin{equation}
\mu_k = \mbox{softmax}(\theta_k, \theta) = \frac{\exp{\theta_k}}{\sum_{k'} \exp{\theta_{k'}} }~.
\end{equation}
The simplex can also be mapped to the sphere with the canonical identification $G(\mu)= 2\sqrt{\mu}=x \in \mathbb{S}^{n-1}$ (see for example \cite{lebanon_guy_riemannian_2015}).

In the case of interest for the present paper~\eqref{eq:condprobmodel} we are considering a submodel of the space $\reals^n$, given by the Span of the columns of $V$
\begin{equation}
x_w = 2\sqrt{\mbox{softmax}(\theta_{w}, \theta))} \;, \;\mbox{where }\; \theta_w = V u_w \;.
\end{equation}
The submanifold of the simplex identified by such model is $\mathbb{P}^{n-1}_{\{V\}}$ and the corresponding submanifold of the sphere is $\mathbb{S}^{n-1}_{\{V\}}=\{x \in \reals^n \; | \; G \circ F (\theta),\, \theta\in \spanof{V}\}$. As can be easily verified the points of this set are also points of $\mathbb{S}^{n-1}$ and thus $\mathbb{S}^{n-1}_{\{V\}}\subset \mathbb{S}^{n-1}$.
The tangent space of the submanifold of the sphere is then calculated by means of the pushforward of the composite mapping $G\circ F$,
\begin{equation}
\tang_x \mathbb{S}^{n-1}_{\{V\}}=\{A \in \reals^n \; | \; A= G_* F_* \dot{\theta},\, \mbox{ with } \dot{\theta}\in \spanof{V}\}
\end{equation}
given the obvious identification $\tang_\theta \spanof{V} = \spanof{V}$. A tangent vector on the subsphere can thus be written as
\begin{equation}
\label{eq:xdot}
\dot{x} = \diag(\mu^{-\frac{1}{2}}) \left(I - \mu \otimes \underline{1}\right) \diag(\mu) \;\! \dot{\theta}
\end{equation}
where $\underline{1}$ is the vector with all ones in $\reals^n$ and the $\diag$ of a vector defines a diagonal matrix whose diagonal corresponds to the vector itself.
Notice that the mapping in Eq.~\ref{eq:xdot} is not full rank. In particular the pushforward $\left(I - \mu \otimes \underline{1}\right)$ has rank $n-1$ (null eigenvalue in the direction of $\mu$), corresponding to the fact that each increase of $\theta$ in the direction $\underline{1}$ ($\dot{\theta} \propto \underline{1}$) does not affect the resulting probability distribution.

\section{$h$-representation}
\label{sec:geometry}
We have seen in Section~\ref{sec:subsimplsphere} how the submodel of the simplex, defined by Eq.~\eqref{eq:condprobmodel}, can be mapped to the sphere. In this section we will briefly recap how this reasoning can be extended to a family of diffeomorphisms parametrized by a parameter alpha~\cite{amari:85,amari|nagaoka:2000,amari_information_2010}.

Given a finite sample space $\Dictionary \ni \chi$ of cardinality $n$, the $n-1$ dimensional simplex $\Delta^{n-1}$ embedded in $\reals^n$ is the set of probability distributions over $\Dictionary$.
We denote with $\simplex^{n-1}$ its interior, i.e. $\simplex^{n-1}=\{p \in \reals^n | \sum_{i} p_i = 1,\, p_i>0 \; \forall i \in \Dictionary \}$. In Information Geometry~\cite{amari:85,amari|nagaoka:2000,amari_information_2016}, the interior of the simplex $\simplex^{n-1}$ is commonly represented as a statistical manifold endowed with the Fisher-Rao metric.
A tangent vector $v \in T_p \simplex^{n-1}$ is defined such that $p + \dot p = q$, for some $q \in \simplex^{n-1}$, which gives the common characterization  $\sum_{i=1}^n \dot p_i = 0$ for tangent vectors of the simplex. The Riemannian metric of $\simplex^{n-1} \subset \reals^n$ is 
\begin{equation}
g(p)_{ij} = \frac{1}{p_i} \delta_{ij}\;.    
\end{equation}
Through the metric, it is possible to compute the normal vector in $p$ to $\mathrm T_p \simplex^{n-1}$ with respect to $g$, which is given by $p$ itself, i.e. $\langle p, v \rangle_{g(p)} = 0$ for all $v \in \mathrm T_p \simplex^{n-1}$.
Let us denote $\reals^n_{(-1)}$ as the ambient space of the simplex with the metric $1/p$.
Let us now consider the family of mappings $h_\alpha : \mathcal M_+ \to \mathcal M_+$ 
called $h$-representation of $p$, given by~\cite{amari|nagaoka:2000,amari_information_2010} 
\begin{equation}
h_{\alpha}(p) = \begin{cases}
 \frac{2}{1-\alpha}  p^{\frac{1-\alpha}{2} }  & \alpha \neq 1 \\
 \log p  & \alpha = 1\;,
 \end{cases}
\end{equation}
with derivative $h'_{\alpha}(p) = p^{-\frac{1 + \alpha}{2}}$,
and inverse
\begin{equation}
h^{-1}_{\alpha}(q) = \begin{cases}
\left(\frac{1-\alpha}{2}  q  \right)^{\frac{2}{1-\alpha}}  & \alpha \neq 1 \\
\exp q  & \alpha = 1\;.
\end{cases}
\end{equation}
For $\alpha=0$, the simplex is mapped on the sphere with the canonical identification $h_0(p)= 2\sqrt{p} \in \mathbb{S}^{n-1}$, while for $\alpha=-1$ the mapping becomes the identity.

Let us call $\simplex^{n-1}_{(\alpha)} = h_\alpha(\simplex^{n-1})$. Let $\dot p \in \mathrm T_p \simplex^{n-1}$ be a tangent vector represented in the basis of the ambient space $\reals^n_{(-1)}$, the pushforward is a linear operator $ h_{\alpha *} : \mathrm T_p \simplex^{n-1} \to \mathrm T_{h_\alpha(p)} \simplex^{n-1}_{(\alpha)}$ defined by
\begin{equation}
\label{eq:pushforward}
(h_{\alpha *})_p \, \dot p = p^{-\frac{1 + \alpha}{2} } \dot p.
\end{equation}

In the following we will express all tangent vectors of $\mathrm T_{h_\alpha(p)} \simplex^{n-1}_{(\alpha)}$ in the basis of the ambient space $\reals^n_{(\alpha)}$.
Let $a,b \in \mathrm T_p \simplex^{n-1}$ be two tangent vectors in $p$, for the isometry condition we have
\begin{equation}
\label{metric}
\langle a, b  \rangle_{g(p)} = \langle h_{\alpha *} \,  a, h_{\alpha *}\,  b  \rangle_{g_\alpha(h_{\alpha}(p))} =
\sum_{i=1}^n \frac{a_i b_i}{p_i} = \sum_{i=1}^n p^{-1 - \alpha} a_i b_i p^{\alpha}_i \;.
\end{equation}
The $h$-representation defines a smooth isometry between $\reals^n_{(-1)}$ and $\reals^n_{(\alpha)}$ which is the ambient space $\reals^n$ with the metric induced by the transformation $h_\alpha$, i.e. from Eq.~\eqref{metric} it follows
\begin{equation}
g_\alpha(h_\alpha(p))_{ij} = \delta_{ij} p_i^\alpha \;.
\end{equation}
In other words, the $g_\alpha$ of $\reals^n_{(\alpha)}$, the ambient space of $\simplex^{n-1}_{(\alpha)}$, is defined in such a way that $h_{\alpha}$ is an isometry.
In the following, to favor a lighter notation, we will use $g_\alpha$ to replace $g_\alpha(h_\alpha(p))$.
This mapping also induces an isometry for the $\alpha$-family of Riemannian manifolds $\simplex^{n-1}_{(\alpha)}$, which implies that geodesics can be mapped between manifolds through $h_{\alpha}$ and its inverse. This has a direct computational implication, indeed for $\alpha=0$, the image of $h_0$ is the sphere endowed with the ambient metric of $\reals^n$, for which metric geodesics corresponds to arcs of great circles.

Following a standard construction in Information Geometry due to Amari~\cite{amari:87dual}, we introduce the $\alpha$-family of connections which are flat in the $h$-representation, i.e., the Christoffel symbols in the $h_\alpha$ coordinates vanish. This family of connections allow the definition of $\alpha$-geodesics between two distributions $p$ and $q$, by
\begin{equation}
\gamma_\alpha(t) = h_\alpha^{-1}\left( t\,h_\alpha(p) + (1-t)\, h_\alpha(q)  \right)\;\;\; \mbox{with } t \in [0,1]\;.
\end{equation}
Notice that, unless 
$\alpha = \pm 1$, the curve $\gamma_\alpha(t) \notin \simplex^{n-1}_{(\alpha)}$, and a normalization $c_\alpha(t)$ is required, with $c_\alpha(t) = \sum_{i=1}^n \gamma_{\alpha,i}(t)$~\cite{amari_information_2016}. Since the $\alpha$-geodetic is not metric, $\gamma_\alpha(t) / c_\alpha(t)$ does not represent the shortest path between $p$ and $q$, instead it corresponds to the curve along which vectors remain parallel with respect to the $\alpha$-connection.
The $\alpha$-connection allows also to define a $\alpha$-logarithmic map $\Log^\alpha_p : h_\alpha(\simplex^{n-1}) \to \mathrm T_{h_\alpha(p)} h_\alpha(\simplex^{n-1})$ as the inverse of the $\alpha$-exponential map, which in $\reals^n_{(\alpha)}$ reads
\begin{equation}
\label{log-map}
\Log^\alpha_p(q) = h_\alpha(q) - h_\alpha(p) \;.
\end{equation}

\begin{figure}[htbp]
	\centering
	\includegraphics[width=0.75\textwidth]{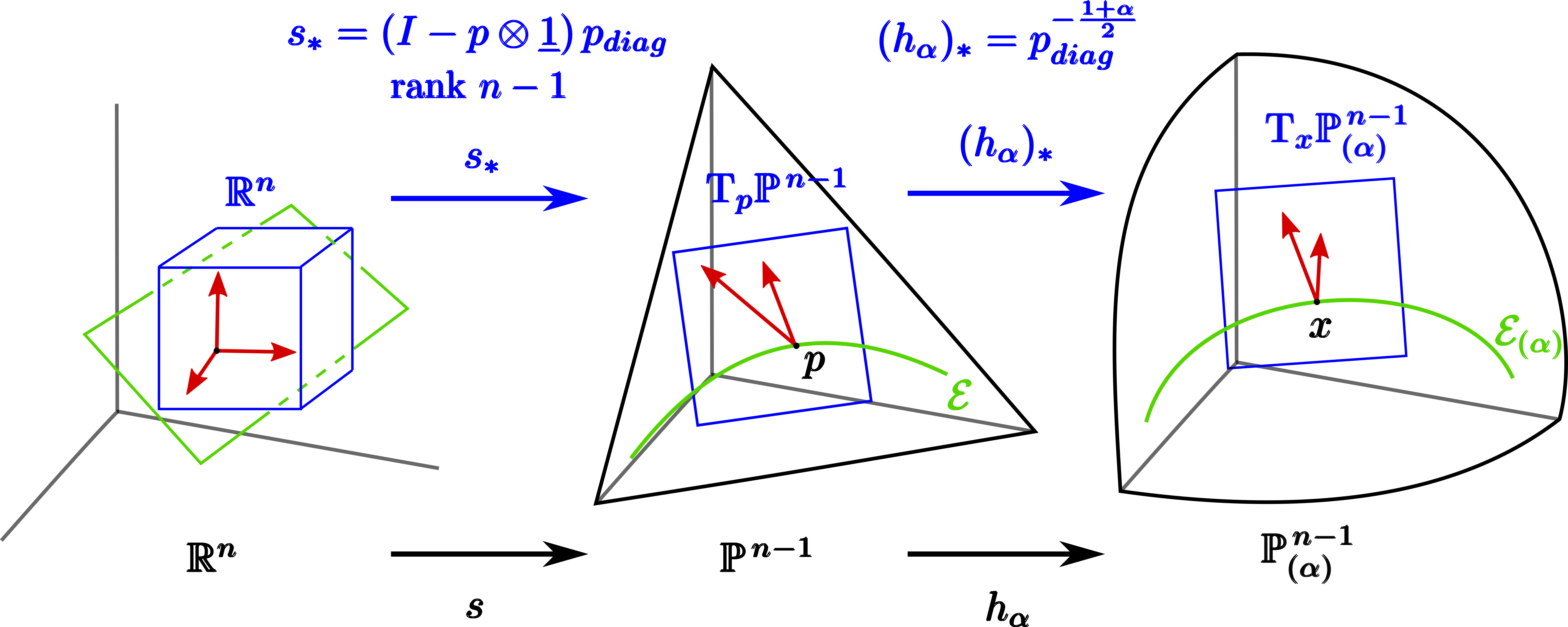}
	\caption{Illustration of the mapping from the over-parametrization of the softmax ($s$) to the $\alpha$-representation of the full simplex and of the low-rank exponential family of this paper $\mathcal{E} \subset \simplex^{n-1}$. Vectors in the tangent space are transported with the pushforward of the composite mapping.}
	\label{fig:submodel}
\end{figure}

Using the typical notation for word embeddings of Section~\ref{sec:exp_model}, with respect to the exponential family $\mathcal E$ in Eq.~\eqref{eq:condprobmodel}, we have that the sample space coincide with the dictionary $\Dictionary$, $u \in \reals^d$ are the natural parameters, and $T(\chi)_k = (V \chi)_k = (v_\chi)_k$ are the sufficient statistics for a given $\chi \in \Dictionary$.
In a slightly different notation, we can rewrite the $d$-dimensional exponential family $\mathcal E \subset \simplex^{n-1}$ of Eq.\eqref{eq:condprobmodel} as
\begin{equation}
\label{p_in_exp_form}
p_u(\chi) = p(\chi; u) =  \exp \left( \, \sum_{k=1}^d u_k V^\chi_k - \psi(u) \right )~,
\end{equation}
where $u = (u_1, \dots, u_d)^\trasp$ is the vector of $d < n$ natural parameters, 
$T(\chi) = (V^\chi_1, \dots, V^\chi_d)^\trasp$ is the vector of sufficient statistics, and 
\begin{equation}
\psi(u) = \log \sum_{\chi} \exp \left(\, \sum_{k=1}^d u_k V^\chi_k \right )
\end{equation}
is the normalizing constant. The tangent space $\mathrm T_{p_u} \mathcal E$ equals $\spanof{ V_k - \mathbb E_{p_u} [V_k]}$, where $V_k$ is a column of the matrix $V$. The Fisher matrix reads
\begin{equation}
 I(u) = \Cov_{p_u}(T, T) = \Cov_{p_u}(V, V) = \sum_\chi p_u(\chi) (V^\chi - \mathbb E_{p_u} [V])\otimes (V^\chi - \mathbb E_{p_u} [V]) ~.   
\end{equation}
The mapping $u \mapsto p$ has a Jacobian given by 
\begin{equation}
\diag(p)\,\Delta V(u)~,     
\end{equation}
where $\Delta V(u) = (V - \mathbb E_{p_u}[V] )$ is the matrix of the centered sufficient statistics.
By applying the $h_\alpha$ mapping to $p \in \mathcal E$ we can map the exponential family in Eq.~\eqref{p_in_exp_form} to a submanifold  $\mathcal{E}_{(\alpha)} \subset h_\alpha(\simplex^{n-1})$. By combining the Jacobian of $u \to p$ with the pushforward in Eq.~\eqref{eq:pushforward},
we obtain a characterization of the tangent space of $\mathcal{E}_{(\alpha)}$ as a linear subspace $\mathrm T_{h_\alpha(p_u)} \mathcal{E}_{(\alpha)} \subset \reals^n_{(\alpha)}$, by
\begin{equation}
	\diag \left(p^\frac{1-\alpha}{2}\right) \Delta V(u) \, \dot u\;,
\end{equation}
where $\dot u$ is a tangent vector in the parameter space $\reals^d$.
See Fig.~\ref{fig:submodel} for a graphical representation.
Through the characterization of the basis, we can calculate the projection $\Pi^\alpha_{u}$ onto the tangent space $\mathrm T_{h_\alpha(p_u)} \mathcal{E}_{(\alpha)}$ of a vector $a \in \reals^n_{(\alpha)}$.
Let us define the matrix 
$A_\alpha(u) = \diag \left(p^\frac{1-\alpha}{2}\right) \Delta V(u) $ whose columns correspond to the basis vectors of $\mathrm T_{h_\alpha(p_u)} \mathcal{E}_{(\alpha)}$ expressed through the basis of the ambient space $\reals^n_\alpha$. The coordinates of the projection of a vector $a$ on the basis $A_\alpha$ of $\mathrm T_{h_\alpha(p_u)} \mathcal{E}_{(\alpha)}$ are
\begin{equation}
\label{projection-formula}
\Pi^\alpha_p (a) =  (A_\alpha^\trasp g_\alpha A_\alpha)^{-1} A_\alpha^\trasp g_\alpha a = I(u)^{-1} \Delta V(u) ^\trasp \left(p^{\frac{1+\alpha}{2}} \odot a \right)~.
\end{equation}

Let us now take two vectors in the ambient tangent space $a, b \in \reals^n_{(\alpha)}$, the inner product in $\mathrm T_{h_\alpha(p_u)} \mathcal{E}_{(\alpha)}$ of their projections reads
\begin{equation}
\label{alpha-similiarty}
\langle \Pi^\alpha_p (a), \Pi^\alpha_p (b) \rangle_{I(u)} = \left(p^{\frac{1+\alpha}{2}} \odot a \right)^\trasp   \Delta V(u) I(u)^{-1} \Delta V(u)^\trasp \left(p^{\frac{1+\alpha}{2}} \odot b \right)~,
\end{equation}
where we use the Fisher metric $I(u)$ in the inner product, since $\Pi^\alpha_p (a)$ is projecting on the basis $A_\alpha$ of the sufficient statistics $\alpha$. We prove the following theorem.
\begin{thm}
\label{thm1}
Let $\alpha = 1$, then Eq.~\eqref{alpha-similiarty} reduces to $\langle u_a-u, u_b-u \rangle_{I(u)}$.
\end{thm}

Let us consider $\alpha=1$, given the distribution of Eq.~\eqref{p_in_exp_form} the Log map for a word $a\in \Dictionary$ becomes
\begin{align}
\Log^{1}_u a &= h_{1}(p_a) - h_{1}(p_u) =  V (u_a-u) - \ln \left(\frac{Z_a}{Z_u}\right) \, \underline{1}~,
\end{align}
where $\underline{1}$ is the vector of all ones. The vector of ones $\underline{1}$ is orthogonal to $p$ in the simplex, that is, it is proportional to the pushforward of the normal vector $p$ to $\simplex^{n-1}$ in $p$ for $\alpha = 1$. The projection of such vector on the tangent space project away the `one' component
\begin{equation}
\begin{split}
\Pi^{1}_{p_u} &(\Log^{1}_u a)^k \\
&= \left( I(u)^{-1} \right)^{ki} \sum_\chi p_u(\chi) (V^\chi_i - \E_p [V^\cdot_i]) \left(V^\chi_j (u_a-u)^j - \ln \left(\frac{Z_a}{Z_u}\right) \, \underline{1}^\chi \right)\\
&= \left( I(u)^{-1} \right)^{ki} \sum_\chi p_u(\chi) (V^\chi_i - \E_{p_u}[V^\cdot_i]) (V^\chi_j  - \E_{p_u} [V^\cdot_j]) (u_a-u)^j \\
&= (u_a-u)^j \;,
\end{split}
\end{equation}
where $i,j,k$ are indices and we used Einstein summation convention whenever possible (paired indices).
We just proved that for $\alpha = 1$, the projection~\eqref{projection-formula} reduces to $u_a-u$.
It follows that, Eq.~\eqref{alpha-similiarty} reduces to $\langle u_a-u, u_b-u \rangle_{I(u)}$.
\begin{cor}
Let $\alpha = 1$, in the uniform distribution $(u=0)$, Eq.~\eqref{alpha-similiarty} reduces to $\langle u_a, u_b \rangle_{I(0)}$.
\end{cor}

\section{A Geometric Framework for Word Embedding}
\label{sec:geometry_embedding}

In this section we apply the geometric framework defined so far with the purpose of defining a family of $\alpha$-measures for word similarities and word analogies based on Information Geometry.

Given $U,V$, and a reference measure $p_u(\chi)$, we introduce a family of geometric measures of $\alpha$-similarity $\Sim^{\alpha}_{u}(a, b)$ for two words $a, b \in \Dictionary$, by generalizing to the Riemannian case the computation of the cosine product between two tangent vectors. The intuition behind this definition is to provide a similarity measure based on the cosine product between two directions in the tangent space of $p_u(\chi)$ pointing towards $p_{a}(\chi) = p_{u_{a}}(\chi)$ and $p_{b}(\chi) = p_{u_{b}}(\chi)$, respectively. The inner products are computed with respect to the Fisher metric and the logarithmic maps with respect to the $\alpha$-connection. The computation of this quantity can be done by first obtaining the $h_{\alpha}$-representations of $p_{a}(\chi)$ and $p_{b}(\chi)$. The second step consists in computing the $\alpha$-logarithm map centered in $p_u$ of these two points using Eq.~\eqref{log-map}, followed by a projection $\Pi^\alpha_{u}$ on the tangent space $T_{h_\alpha(p_u)} h_\alpha( \mathcal E)$
\begin{equation}
\label{compute}
\widehat{\Log^\alpha_{p_u}}(p_w) = \Pi^\alpha_{p_u}( h_\alpha(p_w(\chi)) - h_\alpha(p_u(\chi)))
\end{equation}
see also Eq.~\eqref{projection-formula}. Finally, the last step consists in the evaluation of the Riemannian cosine product with respect to the Fisher matrix.
\begin{definition}
The $\alpha$-cosine similarity between two words $a$ and $b$ with respect to a reference distribution $p_u$ reads
\begin{equation}
\label{sim-index}
\Sim^{\alpha}_{p_u}(a, b) = \frac{\langle \widehat{\Log^\alpha_{p}}(p_{a}),
\widehat{\Log^\alpha_{p}}(p_{b})\rangle_{I(p_u)}}
{|| \widehat{\Log^\alpha_{p}}(p_{a})||_{I(p_u)} \;  ||\widehat{\Log^\alpha_{p}}(p_{b})||_{I(p_u)}}~.
\end{equation}
\end{definition}

For $\alpha=1$, Eq.~\eqref{sim-index} simplifies as stated in the following proposition, based on Theorem~\ref{thm1}.
\begin{proposition}
\label{prop:alpha1uniformeqtoucosprod}
The $\alpha$-cosine similarity between two words $a$ and $b$ with respect to a reference distribution $p$ for $\alpha=1$ simplifies as
\begin{equation}
\Sim^{1}_{p_u}(a, b) = \frac{u_{a}^\trasp I(p_u) u_{b}}{||u_{a}||_{I(p_u)} \; ||u_{b}||_{I(p_u)}}~.
\end{equation} 
\end{proposition}
It is a common approach in the literature of word embeddings to measure the similarity between two words using the cosine product between the embedding vectors $u_{a}$ and $u_{b}$, 
\begin{equation}
\label{standard-cosine}
\Sim(a, b) = \frac{u_{a}^\trasp  u_{b}}{||u_{a}||_2 \; ||u_{b}||_2}~,
\end{equation}
see for example~\cite{pennington_glove:_2014,baroni_dont_2014,bakarov_survey_2018}.
In the light of the previous proposition, the cosine product between the $u$ vectors~\eqref{standard-cosine} is a special case of Eq.~\eqref{sim-index} for $\alpha=1$ and when $I(p_u)$ is isotropic.
The following proposition provides a sufficient condition which guarantees the isotropy of the Fisher information matrix.
\begin{prop}
\label{prop:isotropy}
Let $p_0$ be the uniform distribution, if the sufficient statistics are centered, i.e., $\mathbb E_{p_0} [ v_{\chi}]=0$, and the matrix $V^\trasp V$ is isotropic, then Fisher information matrix $I(p_0)$ is isotropic too.
\end{prop}
Notice that equivalently the standard cosine product in Eq.~\eqref{standard-cosine} corresponds to the case when the computations are done in the natural parameters of the exponential family and the Fisher-Rao metric is replaced by the standard Euclidean metric.

Also the resolution of analogies can be generalized in a geometric way. Given an analogy of the form $a:b = c:d$, we compute the $\alpha$-logarithmic maps $\Log^{\alpha}_{p_a} p_b$ and $\Log^{\alpha}_{p_c} p_d$ and $\alpha$-parallely transport them in the same reference point $p_u$ with the $\alpha$-connection. Since the $\alpha$ connection is flat in $\reals^n_{(\alpha)}$, this simply corresponds to a translation of the vectors. Once in $h_\alpha(p_u)$ the vectors can be projected onto $\mathrm T_{h_\alpha(p_u)} \mathcal{E}_{(\alpha)}$, and the norm of the difference can be computed with respect to the metric. This gives a novel measure of word analogy which depends on $\alpha$ and can be written as
\begin{equation}
{\footnotesize
	\label{eq:analogypenalty}
	    \kappa^{(\alpha)}_{p_u}(p_a, p_b, p_c, p_d) = \left\| \Pi^{\alpha}_{p_u} \left( h_\alpha(p_b(\chi)) - h_\alpha(p_a(\chi)) - h_\alpha(p_d(\chi)) + h_\alpha(p_c(\chi)) \right) \right\|_{I(p_u)} \;.
}
\end{equation}
Given a reference measure $p_u$ and a value of $\alpha$, this quantity can be used to solve an analogy, for instance by minimizing it over $c \in \Dictionary$, i.e., 
	\begin{equation}
	\label{eq:alphaanalogy}
	\argmin_c \kappa^{(\alpha)}_{p_u}(p_a, p_b, p_c, p_d)\;.
	\end{equation}
Let us notice that for $\alpha=1$, and for the exponential family of Eq.~\eqref{p_in_exp_form}, Eq.~\eqref{eq:analogypenalty} reduces to the norm of a vector in the tangent space of $p_u$
\begin{equation}
\kappa^{(1)}_{p_u}(p_a, p_b, p_c, p_d) = \left\| u_b - u_a - u_d + u_c \right\|_{I(p_u)} \;.
\end{equation}
Furthermore, under the conditions in Proposition~\ref{prop:isotropy} which guarantees the isotropy of the Fisher information matrix, we recover the standard formulation of word analogy in Eq.~\eqref{eq:argminrelvectors}.

\section{Alpha Embeddings}
\label{sec:alphaembeddings}
As an alternative view, we propose to use as $\alpha$ embeddings the coordinates of the projected Logarithmic map
\begin{equation}
\label{eq:alpha-emb}
\begin{split}
W^\alpha_{u}(w)^k
& = \Pi_u^\alpha \left( \Log^{\alpha}_{p_u} p_w \right)^k \\ &= \left(I(u)^{-1}\right)^{kj} \sum_\chi l^\alpha_{uw}(\chi) \; \Delta V(u)_j^\chi
\end{split}\;,
\end{equation}
with $k=1, \dots, d$, and where the coefficients
\begin{equation}
\label{eq:alpha-ldv}
l^\alpha_{uw}(\chi) =
\begin{cases}
p_u(\chi) (\ln p_w(\chi) - \ln p_u(\chi)) \qquad &\alpha = 1\\[10pt]
p_u(\chi) \frac{2}{1-\alpha} \left( \left(\frac{p_w(\chi)}{p_u(\chi)}\right)^\frac{1-\alpha}{2} - 1 \right) \qquad &\alpha\ne 1
\end{cases}
\end{equation}
are changing with alpha and represents the weights of a linear combination of the rows of the matrix of the centered sufficient statistics $\Delta V$.
While the alpha embeddings of Eq.~\eqref{eq:alpha-emb} can be computed in any point $p_u$ and for any value of $\alpha$, they conveniently reduce to $u_w$ in the uniform distribution and for $\alpha=1$, due to Theorem~\ref{thm1}.

Fixed $U, V$ and a reference point $p_u$ we can thus rewrite similarity and analogy measure in terms of the alpha embeddings as
\begin{equation}
\label{eq:alpha-similarity}
\Sim^\alpha(a,b) = \frac{\langle W^\alpha_{u}(a), W^\alpha_{u}(b) \rangle_{I(p_u)}}{|| W^\alpha_{u}(a) ||_{I(p_u)} ||W^\alpha_{u}(b)||_{I(p_u)}}
\end{equation}
and
\begin{equation}
\kappa^{(\alpha)}_{p_u}(p_a, p_b, p_c, p_d) = \left\|
W^\alpha_{u}(b) - W^\alpha_{u}(a) - W^\alpha_{u}(d) + W^\alpha_{u}(c)
\right\|_{I(p_u)}\;.
\end{equation}

\section{Limit Embeddings}
Let us notice that: when $\alpha \to +\infty$, the weight factor $l^\alpha_{uw}(\chi)$ goes to zero for all words $\chi$ such that $p_w(\chi) > p_u(\chi)$ and grows on the others.
Vice versa for $\alpha \to -\infty$, the weight factor grows for all words $\chi$ such that $p_w(\chi) > p_u(\chi)$ and goes to zero for the others.
The limit of $\alpha \to -\infty$ for a word $w$ then becomes interesting, since the factors $l^\alpha_{uw}(\chi)$ will tend to a delta on the word $\chi_w^* = \argmax_\chi p_{w}(\chi)/p_{u}(\chi)$. To achieve a numerically stable version of these embeddings we propose the following formula
\begin{equation}
\label{eq:limit-emb}
\begin{split}
LE^\alpha_{u}(w)^i
& = \left(I(u)^{-1}\right)^{ij} \; \Delta V(u)_j^{\chi^*_w}\;,
\end{split}
\end{equation}
we will call this the limit embedding (LE), which is depending only to the $\chi^*_w$ row of the matrix of sufficient statistics $\Delta V(u_0)$.
This leads to very simple geometrical evaluation tasks of similarities and analogies in the limit.

\section{Computational Stability}
For negative alphas, the formula \eqref{eq:alpha-emb} gets numerically unstable pretty quickly. Since the interest is usually on the directions (cosine product on similarities and vectors are often normalized before evaluating analogies) we propose to compute a numerically stable version of $l^\alpha_{0w}(\chi)$ as 
\begin{equation}
\label{eq:alpha-ldv-scaled}
\small{
\begin{split}
\tilde{l}^\alpha_{0w}(\chi) 
= p_0(\chi) \left( \left(\frac{\frac{p_w(\chi)}{p_0(\chi)}}{\|\frac{p_w(\chi)}{p_0(\chi)}\|_\infty} \right)^\frac{1-\alpha}{2} - \frac{1}{\|\frac{p_w(\chi)}{p_0(\chi)}\|_\infty^\frac{1-\alpha}{2}} \right)
\end{split}
}
\end{equation}
which is the rescaled version of \eqref{eq:alpha-emb}, up to a normalization factor independent on $\chi$. We will use this normalization trick to obtain numerically stable alpha embedding vectors in the rest of the paper.

\section{Change of reference measure}
\label{sec:changeref}
In the literature of word embedding it is common practice to consider $u_w+v_w$ as embedding vectors for calculating similarities and analogies~ \cite{bullinaria_extracting_2012,mikolov_efficient_2013,mikolov_distributed_2013,pennington_glove:_2014,raunak_simple_2017}.
The embedding vectors given by the sum $u+v$ have been experimentally shown to provide better results~\cite{pennington_glove:_2014} compared to using simply $u$.
With regard to Equation~\eqref{eq:condprobmodel}, summing $u$ and $v$ vectors, corresponds to a shifting of the natural parameters $u$ of the exponential family.
For each word $w$ this shift is different and it can be interpreted as a change of reference measure for the conditional distribution of that particular word. The reweighted probabilities are
\begin{equation}
\label{eq:changeref}
\begin{split}
p^{(+)}(\chi | w) &= \frac{\exp\left(\left(u_w+v_w\right)^\trasp v_\chi\right)}{Z^{(+)}_w} \\
&= p(\chi | w) \frac{Z_w \exp\left( v_w^\trasp v_\chi \right)}{Z^{(+)}_w} = p(\chi | w) \, r(\chi; w)
\end{split}
\end{equation}
where $Z^{(+)}_w$ is the partition function for the $u+v$ vectors, and $r(\chi; w)$ is the reference measure used for the word $w$.
The reference measure $r(\chi; w)$ is based on the scalar product between the outer vectors $v$ (which are interpreted as the sufficient statistics, see Sec.~\ref{sec:exp_model} and \ref{sec:geometry}). $r(\chi; w)$ is higher for those words which behaves more similarly to the word itself when in the context (similar direction for the outer vectors). Using Equation~\eqref{eq:changeref}, in place of Eq.~\eqref{eq:condprobmodel} as starting point to calculate the alpha embeddings, we obtain the U+V alpha embeddings.

\section{Model Training}
We have performed experiments using the English Wikipedia dump from October 2017 (enwiki).
We used the wikiextractor python script~\cite{attardietal2017} to parse the Wikipedia dump xml file. We decide to use a simple preprocessing to have a standard baseline: we lower case all the letters, we remove stop-words and we remove punctuation. To obtain the dictionary of words for the enwiki corpus we use a cut-off minimum frequency (m0) of 1000. The words occurring less than m0 times in the corpus are agglomerated in a single unknown token ($unk$). In this way we obtained a dictionary of 67,336 words. In accordance with~\cite{pennington_glove:_2014} we choose a window size of 10 around each word (10 words preceding and 10 following) with decaying weighting rate from the center of $1/d$ for cooccurrences calculation. We trained our models with Glove~\cite{pennington_glove:_2014} with vector sizes of 100, 200 and 300, for a maximum of 1000 epochs (each epoch means iterating over all the entries of the cooccurrence matrix).
To make sure that the models trained are effectively comparable with the models in the literature we evaluate accuracies on the word analogy tasks of \cite{pennington_glove:_2014,mikolov_distributed_2013,mikolov_efficient_2013}. We decide to keep the vectors obtained after 1000 epochs since the training has converged. To verify the correct convergence of the training we tested on the analogy tasks of \cite{pennington_glove:_2014,mikolov_distributed_2013,mikolov_efficient_2013} using the code available from the GloVe paper~\cite{pennington_glove:_2014}. We also analyzed the performances of the model in similarity during training as we will see more in details in the Results section.
\begin{table}[htbp]
	\begin{tabular}{|c|ccccc|}
\hline
		corpus & vec size & iter & Sem. & Syn. & Tot. \\ \hline
        \multirow{15}{*}{enwiki 1.48B} & \multirow{5}{*}{100}
                    & 200  & 67.40  & 55.11  & 60.39  \\
                &	& 400  & 69.13  & 55.40  & 61.30  \\
                &	& 600  & 69.38  & 55.51  & 61.47  \\
                &	& 800  & 69.72  & 55.51  & 61.62  \\
                &	& 1000 & 69.85  & 55.47  & 61.65  \\ \cline{2-6} & \multirow{5}{*}{200}
                    & 200  & 77.38  & 62.14  & 68.69  \\
                &	& 400  & 78.22  & 62.65  & 69.35  \\
                &	& 600  & 78.56  & 62.52  & 69.42  \\
                &	& 800  & 78.83  & 62.62  & 69.59  \\
                &	& 1000 & 78.99  & 62.74  & 69.72  \\ \cline{2-6} & \multirow{5}{*}{300}
                    & 200  & 80.79  & 63.83  & 71.12  \\
                &   & 400  & 82.21  & 64.32  & 72.01  \\
                &   & 600  & 82.46  & 64.53  & 72.24  \\
                &   & 800  & 82.54  & 64.60  & 72.31  \\
                &   & 1000 & 82.54  & 64.66  & 72.34  \\
        \hline
        \multirow{2}{*}{enwiki 1.6B GloVe paper\cite{pennington_glove:_2014}}
            & 100 & 50   & 67.5   & 54.3   & 60.3   \\ \cline{2-6}
            & 300 & 100  & 80.8   & 61.5   & 70.3   \\
\hline
	\end{tabular}
	\caption{Accuracy on the word analogy tasks of \cite{pennington_glove:_2014,mikolov_distributed_2013,mikolov_efficient_2013}}
	\label{tab:wordanalogies}
\end{table}

\section{Results}
As discussed in Section~\ref{sec:alphaembeddings} the alpha embeddings (denoted by E in tables and figures) can be calculated in any point on the manifold. In this section we consider three points of interest: the uniform distribution (0), the unigram distribution from the model (u) obtained calculating the marginals of the joint model deriving from Eq.~\ref{eq:condprobmodel} for fixed $U$ and $V$, and the unigram distribution as obtained from the cooccurrences in the data (ud). The vectors are calculated for U or for U+V, with the change of reference measure explained in Section~\ref{sec:changeref}.
\begin{figure}
\centering
\begin{tabular}{cc}
\includegraphics[width=0.5\textwidth]{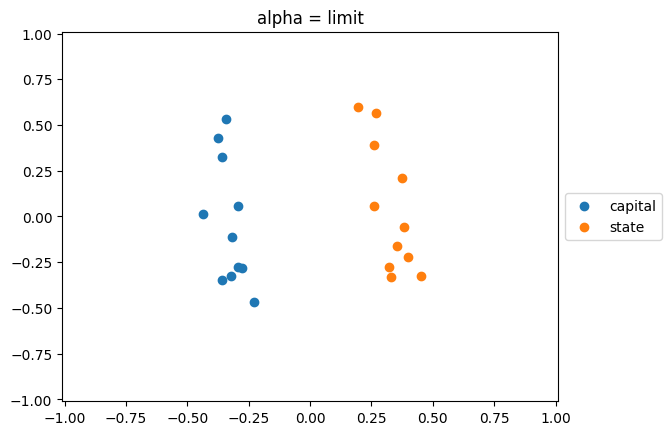} & 
\includegraphics[width=0.5\textwidth]{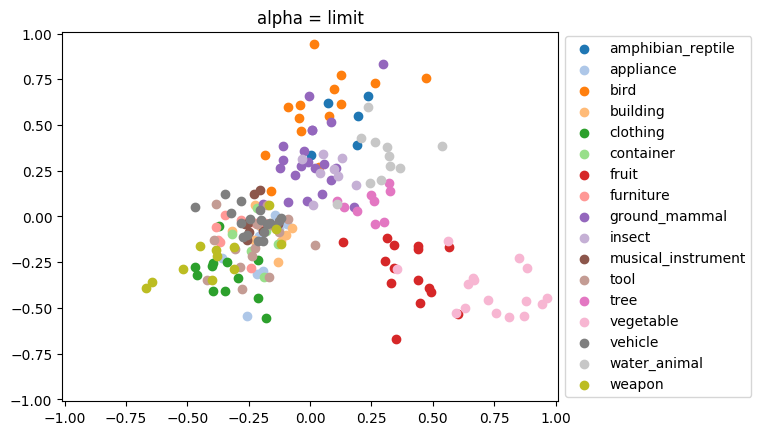} \\
\includegraphics[width=0.5\textwidth]{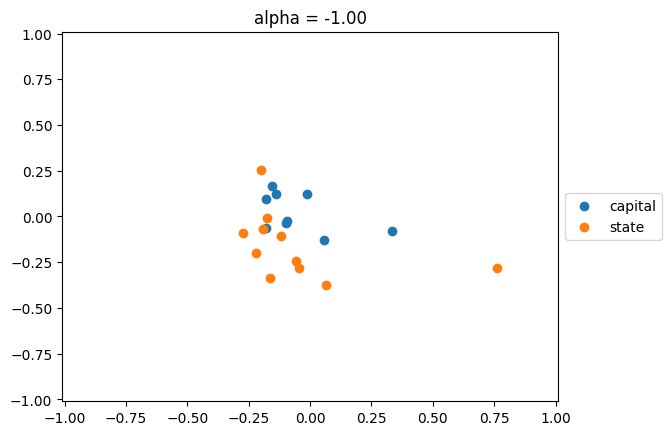} & 
\includegraphics[width=0.5\textwidth]{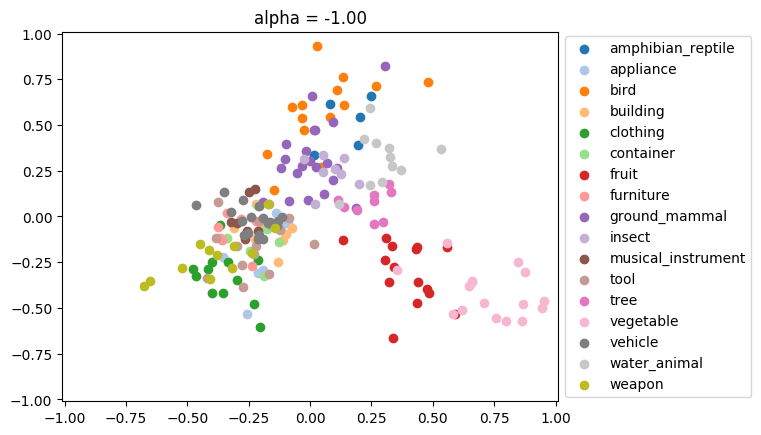}\\
\includegraphics[width=0.5\textwidth]{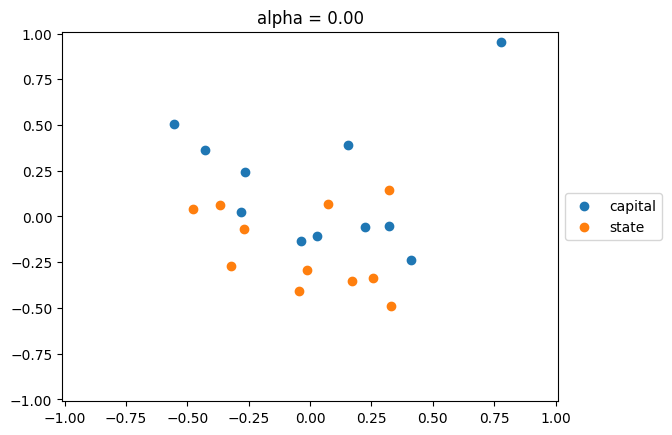} & 
\includegraphics[width=0.5\textwidth]{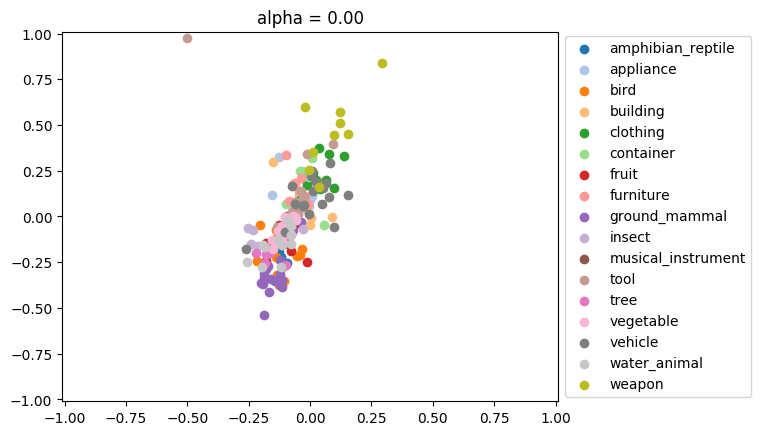}\\
\includegraphics[width=0.5\textwidth]{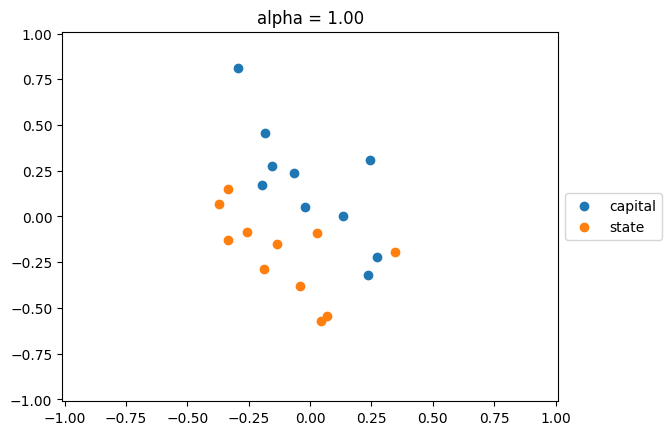} & 
\includegraphics[width=0.5\textwidth]{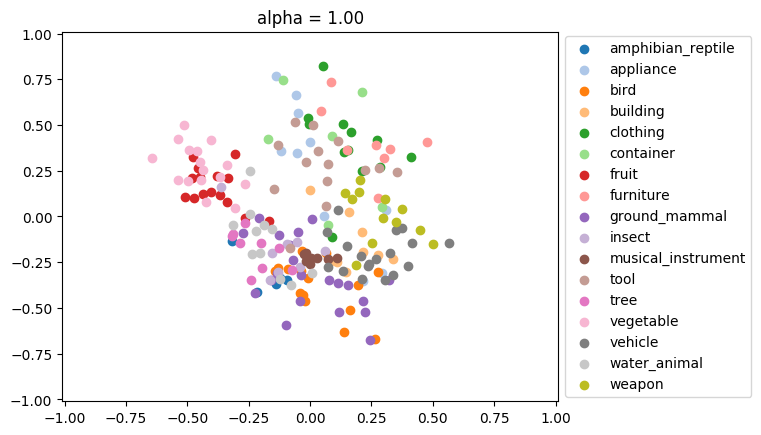}\\
\includegraphics[width=0.5\textwidth]{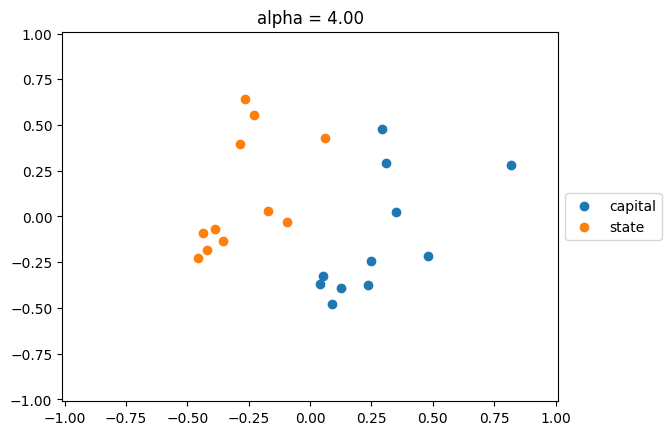} & 
\includegraphics[width=0.5\textwidth]{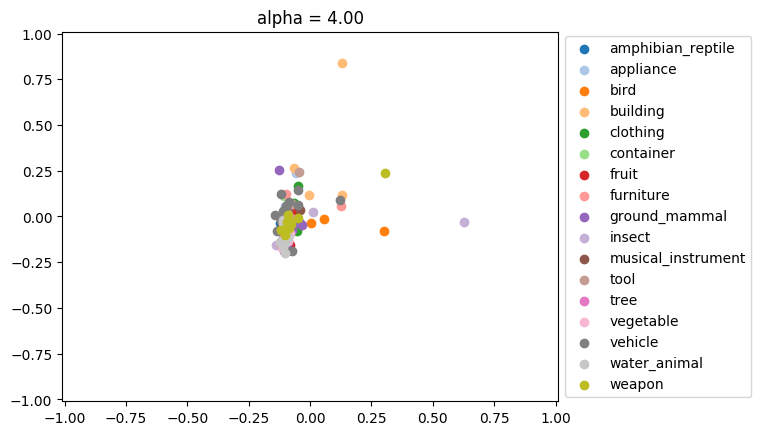}\\
\end{tabular}
\caption{\label{fig:pcaemb}
PCA2D of the embedding vectors for different values of alpha. On the left we visualize two simple groups made of capitals and their respective states (same as Figure~2 of~\cite{mikolov_distributed_2013}). On the right we use the concept category dataset BLESS~\cite{baroni2011we}. The shown PCA are of embeddings in 0 for capitals and in ud for BLESS.}
\end{figure}
\begin{figure}[htbp]
	\centering
	\includegraphics[width=\textwidth]{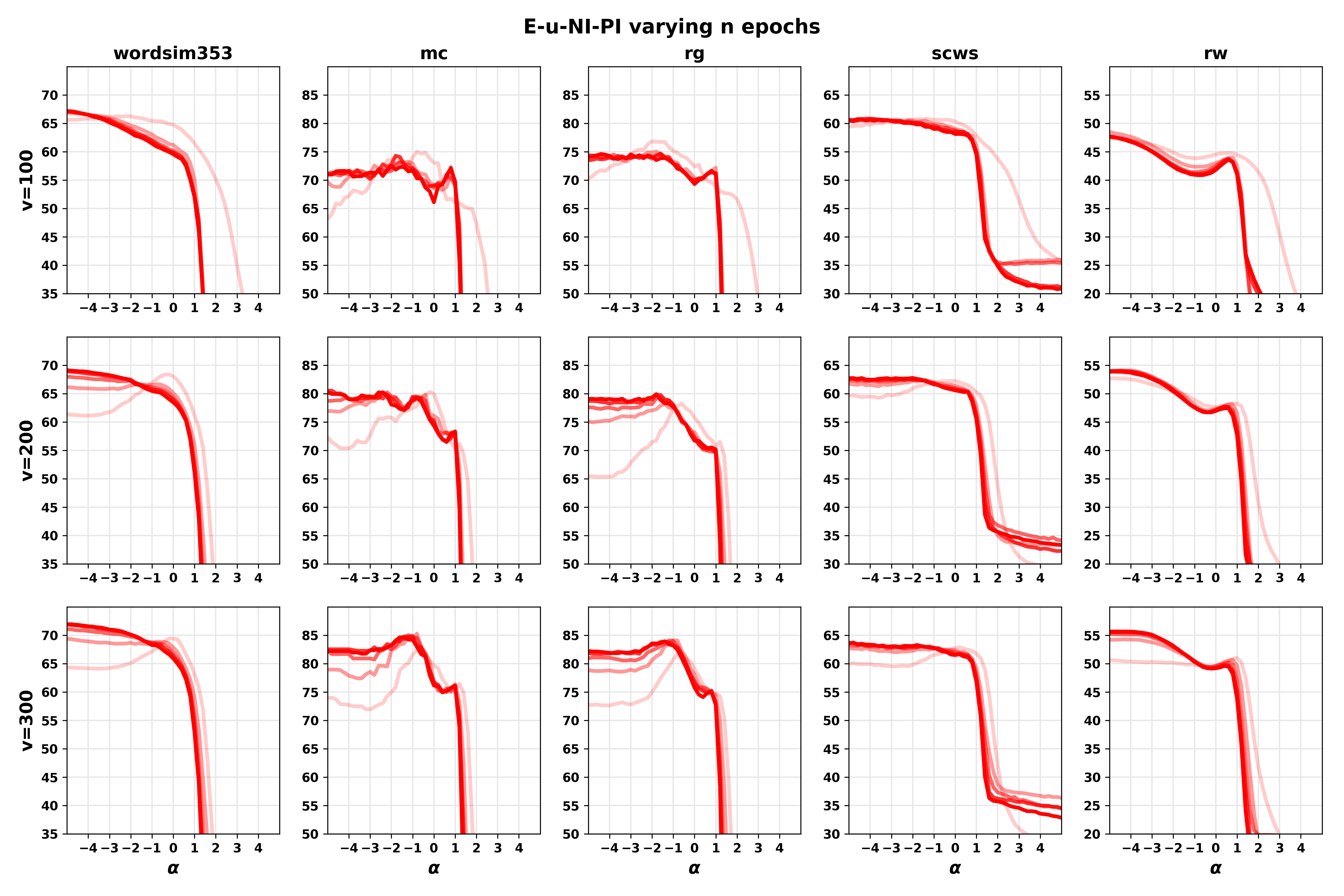}
	\caption{E-u-NI-PI similarities on enwiki, in each row we present models with the same vector size, while in each figure different curves refers to different training epochs, lighter in the beginning of training, the stronger color is at the end of training at 1000 epochs.}
	\label{fig:E-u-NI-PI-vn}
\end{figure}
\begin{figure}[htbp]
	\centering
	\includegraphics[width=\textwidth]{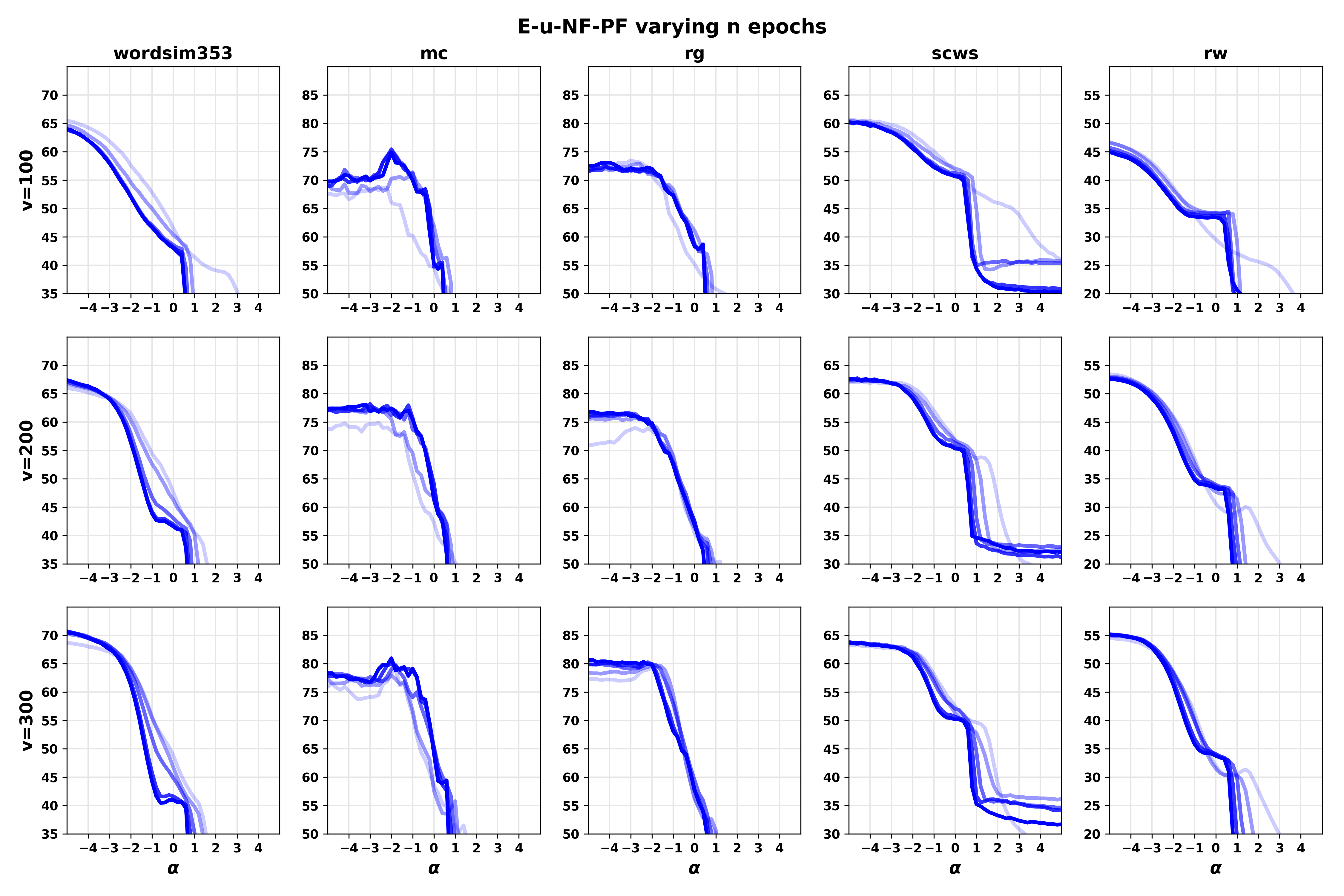}
	\caption{E-u-NF-PF similarities on enwiki, in each row we present models with the same vector size, while in each figure different curves refers to different training epochs, lighter in the beginning of training, the stronger color is at the end of training at 1000 epochs.}
	\label{fig:E-u-NF-PF-vn}
\end{figure}
\begin{figure}[htbp]
	\centering
	\includegraphics[width=\textwidth]{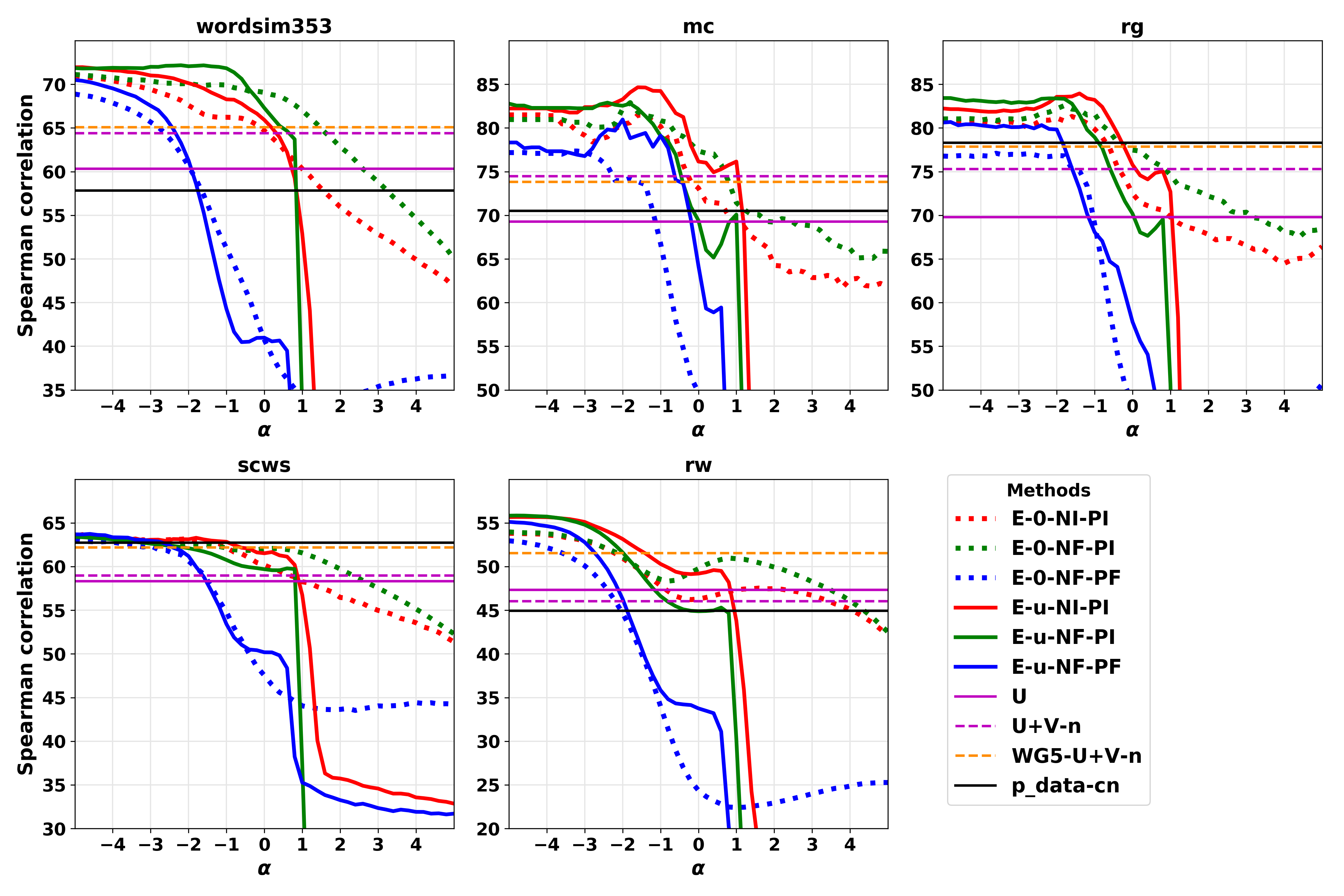}
	\caption{Fully trained model. Similarities for varying alpha, horizontal lines represent baseline methods for reference.}
	\label{fig:similaritiessubmodel}
\end{figure}
\begin{table}[htbp]
	{\footnotesize
		\setlength\tabcolsep{3pt}
		\begin{tabular}{|c|cccccc|}
			\hline
			\rotatebox[origin=c]{90}{\ vecsize\ }    &    method    &    wordsim353    &     mc     &    rg    &   scws     &    rw    \\
			\hline
			\multirow{8}{*}{\rotatebox[origin=c]{90}{100}} 
			&     E-0-NI-PI  &    66.53 (-5.0)  &    71.86 (-2.8)  &    72.98 (-2.2)  &    60.07 (-4.6)  &    44.58 (-4.4)    \\
			&     E-0-NF-PI  &    \bld{67.45 (-5.0)}  &    71.57 (-2.6)  &    72.38 (-1.6)  &    59.87 (0.0)  &    46.08 (1.8)    \\
			&     E-0-NF-PF  &    63.09 (-5.0)  &    68.94 (-3.2)  &    69.41 (-2.8)  &    59.28 (-5.0)  &    41.76 (-5.0)    \\
			&    E-u-NI-PI  &    67.10 (-5.0)  &    \bld{72.50 (-2.0)}  &    \bld{74.60 (-4.4)}  &    \bld{60.72 (-4.4)}  &    \bld{47.72 (-5.0)}    \\
			&    E-u-NF-PI  &    \bld{68.18 (-4.6)}  &    72.08 (-5.0)  &    \bld{74.12 (-3.4)}  &    \bld{60.19 (-5.0)}  &    \bld{47.88 (-5.0)}    \\
			&    E-u-NF-PF  &    64.00 (-5.0)  &    \bld{74.82 (-2.0)}  &    73.08 (-4.0)  &    60.18 (-5.0)  &    44.91 (-5.0)    \\
			&   U  &    60.64       &    64.36       &    67.68       &    57.05       &    43.23         \\
			&    U+V-n  &    62.50       &    68.88       &    70.65       &    57.68       &    42.80         \\
			\hline
			\multirow{8}{*}{\rotatebox[origin=c]{90}{200}}
			&     E-0-NI-PI  &    68.25 (-5.0)  &    79.09 (-3.6)  &    77.67 (-2.6)  &    62.22 (-3.2)  &    52.21 (-4.8)    \\
			&     E-0-NF-PI  &    68.58 (-5.0)  &    79.78 (-5.0)  &    77.94 (-4.8)  &    61.76 (-5.0)  &    52.54 (-5.0)    \\
			&     E-0-NF-PF  &    66.23 (-5.0)  &    75.58 (-5.0)  &    74.80 (-5.0)  &    61.47 (-4.4)  &    50.73 (-5.0)    \\
			&    E-u-NI-PI  &    \bld{69.12 (-5.0)}  &    \bld{80.23 (-5.0)}  &    \bld{79.89 (-1.8)}  &    \bld{62.75 (-2.0)}  &    \bld{53.86 (-4.8)}    \\
			&    E-u-NF-PI  &    \bld{69.38 (-2.6)}  &    \bld{82.21 (-3.6)}  &    \bld{79.63 (-3.6)}  &    62.47 (-5.0)  &    \bld{54.19 (-5.0)}    \\
			&    E-u- NF-PF  &    67.36 (-5.0)  &    78.00 (-3.2)  &    76.82 (-4.8)  &    \bld{62.56 (-4.2)}  &    52.59 (-5.0)    \\
			&    U  &    60.24       &    68.86       &    67.07       &    57.72       &    45.37         \\
			&    U+V-n  &    63.81       &    73.93       &    72.64       &    58.41       &    44.78         \\
			\hline
			\multirow{8}{*}{\rotatebox[origin=c]{90}{300}}
			&     E-0-NI-PI  &    71.01 (-5.0)  &    81.50 (-5.0)  &    81.32 (-1.6)  &    63.55 (-4.4)  &    53.81 (-5.0)    \\
			&     E-0-NF-PI  &    71.12 (-5.0)  &    \bld{82.90 (-1.8)}  &    82.56 (-1.8)  &    63.27 (-5.0)  &    53.97 (-5.0)    \\
			&     E-0-NF-PF  &    68.88 (-5.0)  &    77.36 (-3.4)  &    77.10 (-3.6)  &    63.00 (-5.0)  &    52.95 (-5.0)    \\
			&    E-u-NI-PI  &    \bld{71.96 (-4.8)}  &    \besth{84.66 (-1.6)}  &    \besth{83.95 (-1.4)}  &    \bld{63.72 (-4.8)}  &    \bld{55.67 (-4.8)}    \\
			&    E-u-NF-PI  &    \besth{72.18 (-2.2)}  &    \bld{82.90 (-2.4)}  &    \bld{83.43 (-5.0)}  &    63.40 (-5.0)  &    \besth{55.84 (-4.8)}    \\
			&    E-u-NF-PF  &    70.50 (-5.0)  &    80.96 (-2.0)  &    80.68 (-4.8)  &    \besth{63.74 (-4.6)}  &    55.11 (-5.0)    \\
			&    U           &    60.33       &    69.28       &    69.78       &    58.32       &    47.33         \\
			&    U+V-n       &    64.42       &    74.49       &    75.28       &    58.98       &    46.04         \\
			\hline
			\rotatebox[origin=c]{90}{\ 300\ }    &    WG5-U+V     &    65.08       &    73.82       &    77.85       &    62.18       &    51.54     \\\hline
			-      &    p\_data-cn         &    57.83       &    70.50       &    78.30       &    62.73       &    44.94         \\ \hline
		\end{tabular}
		\caption{enwiki epoch 1000 similarity for the different methods. If the method is dependent on alpha, the best similarity value with the corresponding best alpha value is also reported in parenthesis (scan is between -5 and 5 with step of 0.2 in this table).\label{tab:wordsimilarities-enwiki-epoch1000}
		}}
	\end{table}
In Figure~\ref{fig:pcaemb} we show how the space of word embeddings is deformed with alpha. See also videos in Supporting Information. We will show how this deformation can impact on the evaluation of standard tasks, like for example computing similarity evaluation for varying alpha. Similarity is evaluated by means of the Spearman correlation between the human scores of the dataset and the similarity scores of the method~\cite{mikolov_efficient_2013,pennington_glove:_2014}.
We compare our selected similarity measures with methods from the literature using cosine product on the $u$ vectors (denoted as U) and also cosine product on the vectors $(u+v)/2$ as in~\cite{pennington_glove:_2014} (denoted as U+V). The authors of~\cite{pennington_glove:_2014} report normalizing the columns of the two matrices before the similarity evaluation, and indeed we notice that this process tends to increase the correlations on their methods. This normalization reminds of a Caron factor of 0, cf.~the analysis of Bullinaria et al.~\cite{bullinaria_extracting_2012}, and it is thus linked to the weighting of PCA components in the tangent space, which has been explicitly explored by several authors.~\cite{bullinaria_extracting_2007,bullinaria_extracting_2012,mu_all-but--top:_2017,raunak_simple_2017}
Among the base methods to compare with, we selected a simple method which is a variant of what reported also in~\cite{bullinaria_extracting_2012}, the cosine product of the rows of $p(\chi|w)$ (a row is for fixed $w$), after centering and normalizing its columns (p\_data-cn). Notice that this method does not require training, since the matrix $p(\chi|w)$ can be simply estimated from the cooccurrences~\cite{bullinaria_extracting_2012}.
\begin{figure}[htbp]
	\centering
	\includegraphics[width=0.7\textwidth]{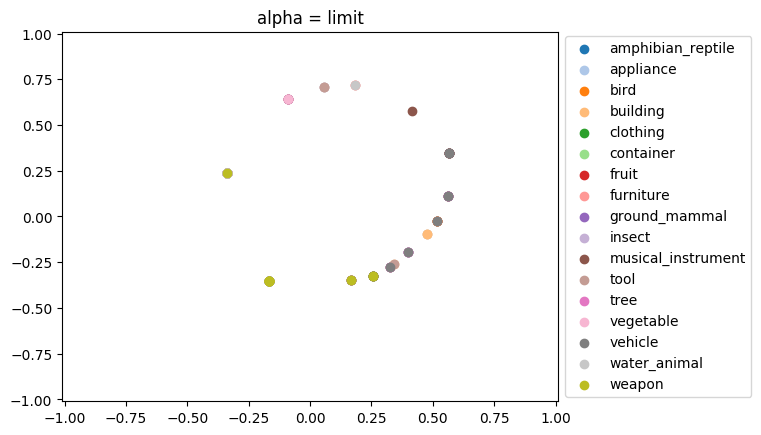}
	\caption{Limit embeddings (LE) from GloVe trained with 2 components, on the word groups of the BLESS dataset.}
	\label{fig:limit2D}
\end{figure}

We test different cosine products in  the tangent space, in which vectors are normalized (N) either with the Fisher matrix (NF) or with the identity (NI), and subsequently the scalar product (P) is performed either with the Fisher matrix (PF) or with the identity (PI).
We have shown in Section~\ref{sec:geometry_embedding} how, in $\alpha=1$, in the point 0 and in case the Fisher is isotropic, Eq.~\eqref{sim-index} reduces to Eq.~\eqref{standard-cosine} commonly used in the literature, see Propositions~\ref{prop:alpha1uniformeqtoucosprod} and \ref{prop:isotropy}. This corresponds to the method E-0-NI-PI which for $\alpha=1$ reduces to the standard scalar product in the Euclidean space U (as can also be observed in the Figures reporting similarities for varying alphas). Analogously, this holds also for the U+V alpha methods, whose dependence on alpha is akin to the U alpha methods. U+V alpha methods are not reported in the plots to not overcrowd them, but they will be further discussed later in the present section. In Figures~\ref{fig:E-u-NI-PI-vn} and \ref{fig:E-u-NF-PF-vn} we show how the similarities of NI-PI and of NF-PF varies during training, for different vector sizes $d$. Notice in the figures how the curves get progressively more flat during training, and simultaneously the similarities tend to improve for very negative alphas. The impact of vector size on the similarity correlations is further detailed in Table~\ref{tab:wordsimilarities-enwiki-epoch1000}. In this table we reported for comparison also WG5-U+V similarities in which we took the online available word embeddings trained on WikiGiga5 corpus~\cite{pennington_glove:_2014}. Let us notice that WG5 vectors are trained on a much bigger corpus, of about 6B tokens. For the sake of a fair comparison, the correlations of the WG5 are computed on the similarities between words belonging to the smaller enwiki dictionary. This in theory constitutes a direct advantage for WG5 since we are restricting to a smaller dictionary made of more frequent (thus supposedly easier) words.

In Figure~\ref{fig:similaritiessubmodel} we plot the alpha methods against the baseline methods from the literature. We can notice how the alpha methods reported in this Figure perform better than the baselines for negative alphas. Methods in ud (not plotted) are found to reach analogous performances of their counterparts in 0 and u, but they decay abruptly for very negative alphas. Remarkably all methods using the limit embeddings of Eq.~\eqref{eq:limit-emb} have good performances, for all points considered: 0, u and ud. LE methods are not reported with horizontal lines in Fig.~\ref{fig:similaritiessubmodel} to not impair the readability of the figure, but their values can be found in Table~\ref{tab:simlevyetal}, and they will be analyzed when comparing with the literature.

The question arises on the origin of the good performances of limit embeddings, or more in general of alpha embeddings for large negative alphas. Let us notice (Eq.~\eqref{eq:limit-emb}) that limit embeddings are performing a clustering in space, in which the same limit embedding vector can be associated to one or more words. We hypothesize that this clustering learned by the system during training (corresponding to the learned sufficient statistics $\Delta V$) is good at extracting the relevant information for similarities. For demonstrative purpose we trained an extra GloVe model with only 2 components. Since the components are only 2 they can be directly plotted and we can see how, in the limit case described in Section~\ref{sec:alphaembeddings}, the limit embeddings LE are indeed corresponding to a form of clustering in space (Figure~\ref{fig:limit2D}). The embeddings have not been normalized in this figure, the rows of the $\Delta V$ matrix resulted having similar norms after the training, in this simple case.
\begin{table}[!bhtp]
	\setlength\tabcolsep{3pt}
	\begin{tabular}{|c|cccccc|}
		\hline
		method &  WSsim  & WSrel  & MEN & MTurk &  RW & SimLex \\
		\hline
		LE-U-0-F & 75.92 & 67.49 & 74.56 & 68.49 & 51.17 & 35.86 \\
		LE-U-0-I & 76.60 & 67.71 & 74.50 & 66.00 & 51.04 & 37.56 \\
		LE-U-u-F & 70.04 & 59.89 & 71.11 & 67.98 & 48.01 & 32.00 \\
		LE-U-u-I & 72.35 & 62.74 & 72.61 & 68.66 & 49.50 & 32.95 \\
		LE-U-ud-F & \bld{77.27} & \besth{69.30} & 75.21 & 60.40 & 52.26 & 37.94 \\
		LE-U-ud-I & 72.81 & 56.79 & 70.93 & 50.73 & 50.89 & 37.12 \\
		\hline
		LE-U+V-0-F & 75.72 & 66.59 & 74.73 & \bld{68.71} & 54.16 & 37.73  \\
		LE-U+V-0-I & 76.46 & 67.12 & 74.76 & 65.94 & \bld{54.82} & \bld{40.05}  \\
		LE-U+V-u-F & 69.62 & 58.20 & 70.95 & 68.31 & 49.55 & 32.87  \\
		LE-U+V-u-I & 71.96 & 61.27 & 72.59 & \besth{68.98} & 51.37 & 34.00  \\
		LE-U+V-ud-F & \besth{77.78} & \bld{69.21} & \besth{75.57} & 60.13 & \besth{55.56} & \besth{41.57}  \\
		LE-U+V-ud-I & 73.62 & 57.21 & 71.28 & 50.31 & 53.83 & \besth{41.57}  \\
		\hline
		Levy et al. 2015 & 74.6 & 64.3 & \bld{75.4} & 61.6 & 26.6 & 37.5 \\
		\hline
	\end{tabular}
	\caption{Similarities obtained on GloVe model trained with vecsize 300 and window size 10. Comparison between the simple limit alpha methods of the present paper and the results reported by Levy et al. 2015\cite{levy2015improving}. The limit methods here reported refer to the cosine product, i.e. both normalization and scalar product are performed with the same matrix, we omit N and P and we simply report either I or F in the name. In each column: light green is the best and dark grey is the second best.
		\label{tab:simlevyetal}
	}
\end{table}

Even though tuning alpha to get the best possible performing geometry on each task is beyond the scope of the current paper, we would still like to be able to compare some of our results with the literature. Let us consider the limit embeddings (LE). This is a simple method which does not require us to perform any cross-validation to tune alpha. We take as a reference comparison a paper by Levy, Goldberg, and Dagan 2015 \cite{levy2015improving} in which they report the best method in cross-validation with a fixed window size of 10 (varying other hyperparameters). In Table~\ref{tab:simlevyetal} we report comparisons on different similarity datasets. We notice how the limit embeddings obtain better or comparable performances on all tasks. The limit methods with the Fisher metric in ud seems to perform better over all, even though it seems to fall short on the MTurk dataset. The methods using $U+V$, the change of reference measure described in Section~\ref{sec:changeref}, seem to provide an improvement expecially on the rare words and simlex datasets. 

\section{Conclusions}
\label{sec:conclusions}
We defined an Information Geometric framework for word embeddings. We introduced a novel family of measures for word similarities and analogies, depending on a deformation parameter $\alpha$, extending common approaches in the literature to their Riemannian counterparts. We evaluated our proposed measures on standard word similarity tasks and showed how our method can outperform previous approaches for a range of values of $\alpha$, recovering existing approaches for $\alpha=1$.
For the enwiki corpus, we obtained a large improvement compared with baselines.
The analysis done so far is orthogonal with respect to the training, it would be of great interest to develop different methodologies to take advantage of the $\alpha$-representation during learning and possibly learn different geometries during learning.
The limit embeddings seems to provide a very simple and effective method, without the need to tune the alpha parameter. The experimental evaluation of $\alpha$ word analogies and of the performances on different downstream tasks will be object of future studies.

\begin{acknowledgements}
The authors are supported by the DeepRiemann project, co-funded by the European Regional Development Fund and the Romanian Government through the Competitiveness Operational Programme 2014-2020, Action 1.1.4, project ID P\textunderscore37\textunderscore714, contract no. 136/27.09.2016.
\end{acknowledgements}

%
%
\clearpage
\bibliographystyle{spphys}       
\bibliography{text,miscellanea,hypersphere,biblio,informationGeometry}

\end{document}